\documentclass[runningheads,a4paper]{llncs}

\usepackage{amssymb}
\usepackage{graphicx}

\usepackage{times}
\usepackage{graphicx}
\usepackage{latexsym}
\usepackage{algorithm}
\usepackage{float}
\usepackage[noend]{algpseudocode}
\usepackage{amssymb}
\usepackage{amsmath} 
\usepackage{amsfonts}
\usepackage{verbatim} 
\usepackage{longtable}
\usepackage{graphicx}
\usepackage{booktabs}
\usepackage{epstopdf}
\usepackage{longtable}
\usepackage{pgfpages}
\usepackage{multirow}
\usepackage{varwidth}
\usepackage{pifont}
\usepackage{url}

\algnotext{EndFor}
\algnotext{EndIf}

\usepackage{array,booktabs,tabularx}
\newcolumntype{C}{>{\centering\arraybackslash}X}

\usepackage{url}
\urldef{\mailsa}\path|{alfred.hofmann, ursula.barth, ingrid.haas, frank.holzwarth,|
\urldef{\mailsb}\path|anna.kramer, leonie.kunz, christine.reiss, nicole.sator,|
\urldef{\mailsc}\path|erika.siebert-cole, peter.strasser, lncs}@springer.com|

\newcommand{\myhappens}{\textsf{\footnotesize happensAt}}
\newcommand{\myholdsAt}{\textsf{\footnotesize holdsAt}}
\newcommand{\myinitiatedAt}{\textsf{\footnotesize initiatedAt}}
\newcommand{\myterminatedAt}{\textsf{\footnotesize terminatedAt}}

\newcommand{\myhappenss}{\textsf{\scriptsize happensAt}}
\newcommand{\myholdsAts}{\textsf{\scriptsize holdsAt}}
\newcommand{\myinitiatedAts}{\textsf{\scriptsize initiatedAt}}
\newcommand{\myterminatedAts}{\textsf{\scriptsize terminatedAt}}
\newcommand{\mynbfs}{\textsf{\scriptsize not}}

\newcommand{\oled}{\textsf{\footnotesize OLED}}

\newcommand{\ec}{\textsf{\footnotesize EC}}

\begin{document}

\mainmatter  

\title{Distributed Online Learning of Event Definitions}

\titlerunning{Distributed Online Learning of Event Definitions}




\author{Nikos Katzouris\inst{1} \and Alexander Artikis\inst{2,1} \and Georgios Paliouras\inst{1} }
\authorrunning{Katzouris et al.} 

\institute{National Center for Scientific Research ``Demokritos'', Athens, Greece\\
\and
Department of Maritime Studies, University of Piraeus, Piraeus, Greece\\
\email{\{nkatz,a.artikis,paliourg\}@iit.demokritos.gr}}

%
%

\maketitle

\begin{abstract}
Logic-based event recognition systems infer occurrences of events in time using a set of event definitions in the form of first-order rules. The Event Calculus is a temporal logic that has been used as a basis in event recognition applications, providing among others, direct connections to machine learning, via Inductive Logic Programming (ILP). \oled \ is a recently proposed ILP system that learns event definitions in the form of Event Calculus theories, in a single pass over a data stream. In this work we present a version of \oled \  that allows for distributed, online learning. We evaluate our approach on a benchmark activity recognition dataset and show that we can significantly reduce training times, exchanging minimal information between processing nodes.
\end{abstract}

\section{Introduction}


Event recognition systems \cite{etzion2010event} process sequences of \emph{simple events}, such as sensor data, and recognize \emph{complex events}, i.e. events that satisfy some pattern. Logic-based systems for event recognition typically use a knowledge base of first-order rules to represent complex event patterns and a reasoning engine to detect such patterns in the incoming data. The Event Calculus (\ec) \cite{kowalski1986logic} has been used as the basis for event recognition systems \cite{artikis2015event}, offering direct connections to machine learning, via Inductive Logic Programming (ILP) \cite{de2008logical}.

Event recognition applications deal with noisy data streams. Methods that learn from such streams typically build a decision model by a single pass over the input \cite{gama2010knowledge}. \oled \ (Online Learning of Event Definitions) \cite{DBLP:journals/tplp/KatzourisAP16} is an ILP system that learns event definitions in the form of \ec \ theories in a single pass over a relational data stream. In this work we present an extension of \oled, that allows for learning a theory in an online and parallel fashion, from disjoint, possibly geographically distributed data streams (it therefore assumes no shared memory). Our approach is based on a simple parallelization scheme of the core \oled \ functionality and it is approapriate for distributed learning. \oled \ learns clauses in top-down manner, by gradually specializing an over-general clause. Its single-pass strategy is based on the Hoeffding bound \cite{hoeffding1963probability}, a tool that allows to build decision models by estimating their quality on a small subset of the  input. \oled \ uses the Hoeffding bound to estimate the quality of candidate specializations of a clause on subsets of the input stream. In the proposed prallelization strategy, clauses are evaluated independently on disjoint data streams and their scores are combined whenever a specialization decision must be made. We present an evaluation of our approach on an a benchmark activity recognition dataset and show that we can significantly reduce training times, exchanging minimal information between processing nodes. This work, therefore, paves the way for relational learning in high-velocity data streams.

The rest of this paper is structured as follows: In Section \ref{sec:backgound} we present some background on the \ec \ and provide a running example. In  Section \ref{sec:oled} we present \oled and in Section \ref{sec:d-oled} we present its distributed version. In Section \ref{sec:experiments} we present our experimental results, while in Section \ref{sec:related-work} we discuss related work. Finally, in Section \ref{sec:final} we discuss some directions for future work and conclude.


\section{Background and Running Example}  
\label{sec:backgound}   


\begin{table}[t]
\scriptsize
\begin{minipage}{\textwidth}
\begin{center}
\begin{tabular}{ll|l}
\hline
\hline
\textbf{Predicate} & \textbf{Predicate Meaning} & \textbf{Axioms} \\
\hline
\myhappenss$(E,T)$ & Event $E$ occurs at time $T$ & $\myholdsAts(F,T+1) \leftarrow$\\
\myinitiatedAts$(F,T)$ & At time $T$ a period of time for & $\qquad  \myinitiatedAts(F,T). \  \  \  \  \  \  \  \  \  \  \ (1)$\\
~ & which fluent $F$ holds is initiated & ~ \\
\myterminatedAts$(F,T)$ & At time $T$ a period of time for & $\myholdsAts(F,T+1) \leftarrow $\\
~ & which fluent $F$ holds is terminated & $\qquad \myholdsAts(F,T), \  \  \  \  \  \  \  \  \  \  \  \  \  \  \ (2)$ \\
\myholdsAts$(F,T)$ & Fluent $F$ holds at time $T$ & $\qquad \mynbfs \  \myterminatedAts(F,T).$ \\
\hline
\hline
\end{tabular}
\end{center}
\end{minipage}
\vspace{0.01cm}
\caption{\footnotesize The basic predicates and domain-independent axioms of the \ec \  dialect.}\label{table:ec}
\end{table}
\normalsize

The Event Calculus (\ec) \cite{kowalski1986logic} is a temporal logic for reasoning about events and their effects. Its ontology consists of \emph{time points} (integer numbers); \emph{fluents}, i.e. properties that have different values in time; and events, i.e. occurrences in time that may alter fluents' values. 
The axioms of the \ec \  incorporate the \emph{common sense law of inertia}, according to which fluents persist over time, unless they are affected by an event. We use a simplified version of the \ec \
that has been shown to suffice for event recognition \cite{artikis2015event}. The basic predicates and its domain-independent axioms are presented in Table \ref{table:ec}. Axiom (1) states that a fluent $F$ holds at time $T$ if it has been initiated at the previous time point, while Axiom (2) states that $F$ continues to hold unless it is terminated. Definitions for \myinitiatedAt/2 and \myterminatedAt/2 predicates are given in an application-specific manner by a set of  \emph{domain-specific} axioms.

We illustrate our approach using the task of activity recognition, as defined in the CAVIAR project\footnote{
\url{http://homepages.inf.ed.ac.uk/rbf/CAVIARDATA1/}}. The CAVIAR dataset consists of videos where actors perform some activities. Manual annotation (performed by the CAVIAR team) provides ground truth for two activity types. The first type corresponds to simple events and consists of knowledge about the activities of a person at a certain video frame/time point, such as \emph{walking}, or \emph{standing still}. The second activity type corresponds to complex events and consists of activities that involve more than one person, e.g. two people \emph{meeting each other}, or \emph{moving together}. The goal is to recognize complex events as combinations of simple events and additional domain knowledge, such as a person's direction and position.

\begin{table}[t]
\scriptsize
\begin{minipage}{\textwidth}
\begin{tabular}{lll}
\hline
\hline
\textbf{(a)} & ~ & \textbf{(b)}\\
\textbf{\underline{Narrative for time 1:}} & \textbf{\underline{Narrative for time 2:}} & \textbf{\underline{Two Domain-specific axioms:}}  \\
\noalign{\smallskip}
$\mathit{\myhappenss(walk(id_1),1)}$ & $\mathit{\myhappenss(walk(id_1),2)}$ & $\myinitiatedAts(moving(X,Y),T) \leftarrow$ \\
$\mathit{\myhappenss(walk(id_2),1)}$ & $\mathit{\myhappenss(walk(id_2),2)}$ & $\qquad  \myhappenss(\mathit{walk(X),T)},$ \\            
$\mathit{\myholdsAts(coords(id_1,201,454),1)}$ & $\mathit{\myholdsAts(coords(id_1,201,454),2)}$ & $\qquad  \myhappenss(\mathit{walk(Y),T)},$\\
$\mathit{\myholdsAts(coords(id_2,230,440),1)}$ & $\mathit{\myholdsAts(coords(id_2,227,440),2)}$ & $\qquad  \mathit{distLessThan(X,Y,25,T)},$\\
$\mathit{\myholdsAts(direction(id_1,270),1)}$ & $\mathit{\myholdsAts(direction(id_1,275),2)}$ &$\qquad  \mathit{dirLessThan(X,Y,45,T).}$ \\
$\mathit{\myholdsAts(direction(id_2,270),1)}$ & $\mathit{\myholdsAts(direction(id_2,278),2)}$ & ~ \\
\noalign{\smallskip}
\textbf{\underline{Annotation for time 1:}} & \textbf{\underline{Annotation for time 2:}} &$\myterminatedAts(move(X,Y),T) \leftarrow$ \\
\mynbfs \  \myholdsAts$(move(id_1,id_2),1)$ & \myholdsAts$(move(id_1,id_2),2)$ & $\quad  \quad  \myhappenss(\mathit{inactive(X),T)},$ \\
~ & ~ & $\quad \quad   \mathit{distMoreThan(X,Y,30,T)}.$\\
\noalign{\smallskip}
\hline
\hline
\end{tabular}
\vspace{0.5\baselineskip}
\end{minipage}
\caption{\footnotesize \textbf{(a)} Example data from activity recognition. E.g., at time point 1 person $\mathit{id_1}$ is \emph{walking}, her $(x,y)$ coordinates are $(201,454)$ and her direction is $270^{\circ}$. The annotation for the same time point states that persons $id_1$ and $id_2$ are not moving together, in contrast to the annotation for time point 2. \textbf{(b)} An example of two domain-specific axioms in the \ec. E.g. the first clause dictates that \emph{moving} of two persons $X$ and $Y$ is initiated at time $T$ if both $X$ and $Y$ are walking at time $T$, their euclidean distance is less than $25$ and their difference in direction is less than $45^{\circ}$. 
}
\label{table:stream1}
\end{table}
\normalsize

Table 2(a) presents some example CAVIAR data, consisting of a narrative of simple events in terms of \myhappens/2, expressing people's short-term activities, and context properties in terms of \myholdsAt/2, denoting people' coordinates and direction. Table 2(a) also shows the annotation of complex events (long-term activities) for each time-point in the narrative. Negated complex events' annotation is obtained via the closed world assumption (although both positive and negated annotation atoms are presented in Table \ref{table:stream1}, to avoid confusion). Table \ref{table:stream1}(b) presents two domain-specific axioms in the \ec.

Our goal is to learn definitions of complex events in terms of initiation and termination conditions, as in Table \ref{table:stream1}(b). In the Learning from Interpretations \cite{blockeel1999scaling} ILP setting that we use in this work, each training example is an interpretation, i.e. a set of true ground atoms, as in Table \ref{table:stream1}(a). Given a set of training interpretations $\mathcal{I}$ and some background theory $B$, which in our case consists of the domain-independent axioms of the \ec, the goal is to find a theory $H$, such that $B\cup H$ \emph{covers} as many $I\in \mathcal{I}$ as possible, where $B\cup H$ covers $I$ when $I$ is a model of $B\cup H$. Although different semantics are possible, in this work a ``model'' is a stable model \cite{gebser2012answer}.

\section{The \oled \ System}
\label{sec:oled}

\oled \ \cite{DBLP:journals/tplp/KatzourisAP16} learns a theory by joining together independently-constructed clauses, each of which is learnt in an online fashion. \oled \ relies on the Hoeffding bound \cite{hoeffding1963probability} to approximate the quality of a clause on the entire input using only a subset of the input. Given a random variable $X$ with range in $[0,1]$ and an observed mean $\overline{X}$ of its values after $n$ independent observations, the Hoeffding Bound states that, with probability $1 - \delta$, the true mean $\hat{X}$ of the variable lies in an interval $(\overline{X} - \epsilon , \overline{X} + \epsilon)$, where $\epsilon = \sqrt{\frac{ln(1/\delta)}{2n}}$. In other words, the true average can be approximated by the observed one with probability $1-\delta$, given an error margin $\epsilon$ that becomes smaller as the number of observations $n$ increases. 

\oled \ learns a clause in a top-down fashion, by specializing it using literals from a bottom clause \cite{de2008logical}. The Hoeffding bound is utilized in the specialization process a follows: Given a clause evaluation function $G$ and some clause $r$, \oled \ evaluates $r$ and all of its candidate specializations on training examples that stream-in. Assume that after $n$ training examples from the input stream, $r_1$ is $r$'s specialization with the highest observed mean $G$-score $\overline{G}$ and $r_2$ is the second-best one, i.e. $\Delta \overline{G} = \overline{G}(r_1)-\overline{G}(r_2) > 0$. Then by the Hoeffding bound we have that for the true mean of the scores' difference $\Delta \hat{G}$ it holds that $\Delta\hat{G} > \Delta\overline{G} - \epsilon \text{, with probability } 1 - \delta$, where $\epsilon = \sqrt{\frac{ln(1/\delta)}{2n}}$. Hence, if $\Delta\overline{G} > \epsilon$ then $\Delta \hat{G} > 0$, implying that  $r_1$ is indeed the best specialization, with probability $1 - \delta$. In order to decide which specialization to select, it thus suffices to accumulate examples from the input stream until $\Delta\overline{G} > \epsilon$. These examples need not be stored or reprocessed. Each example is processed once to extract the necessary statistics for calculating $G$-scores and it is subsequently discarded, thus giving rise to an online (single-pass) clause construction strategy. To ensure that no clause $r$ is replaced by a specialization of lower quality, $r$ itself is also considered as a potential candidate along with its specializations, ensuring that specializing $r$ is a better decision, with probability $1-\delta$, than not specializing it at all. 

The default specialization process follows a \textsf{\footnotesize FOIL}-like, hill-climbing strategy, where a single literal is added to a clause at each specialization step. However, \oled \ supports different specialization strategies as well, e.g. by allowing to simultaneously try all specializations up to a given clause length, or by supporting user-defined, \textsf{\footnotesize TILDE}-like look-ahead specifications \cite{blockeel1998top}.

\begin{figure}[t]
\centering
\includegraphics[width=1\textwidth]{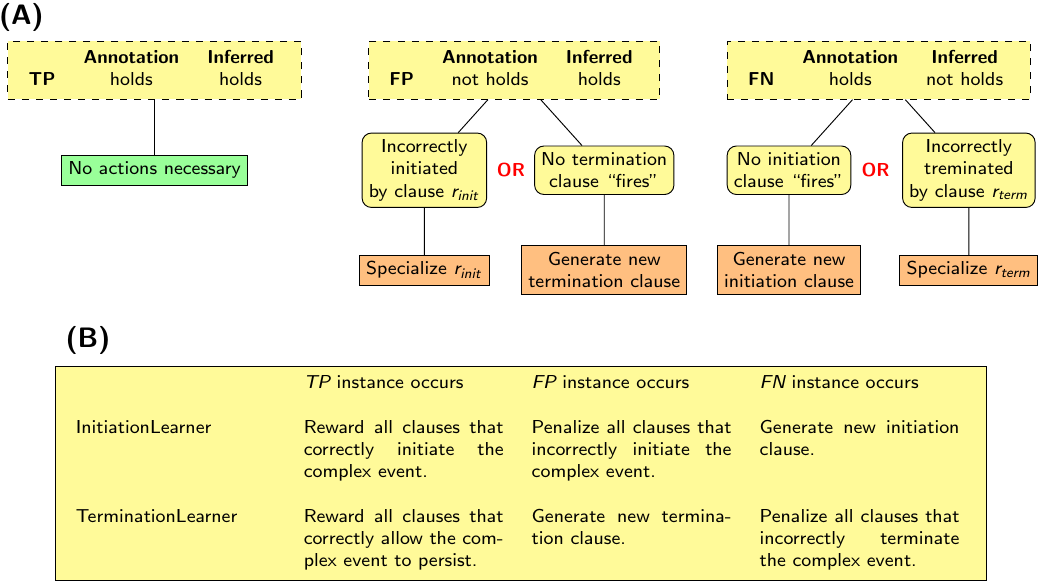}
\caption{\footnotesize \textbf{(A)} Different behaviors of initiation and termination clauses w.r.t. to occurrences of $\mathit{TP, FP}$ and $\mathit{FN}$ complex event instances. Dash-lined boxes explain what it means to encounter a $\mathit{TP, FP, FN}$ complex event instance, in terms of (dis)agreement between the actual label of the instance and the one inferred by the theory. Round-cornered boxes describe the causes of \emph{FP, FN} occurrences w.r.t. the different types of clause (initiation or termination). Regular boxes at the ``leaves'' of the tree-like structures indicate proper courses of action in order to eliminate \emph{FP/FN} instances. \textbf{(B)} Actions taken by the two different processes that learn initiation and termination clauses in parallel, w.r.t. $\mathit{TP, FP, FN}$ complex event occurrences. These actions are in accordance with the indicated actions in \textbf{(A)} (leaves of the trees). ``Rewarding'' a clause refers to increasing the $\mathit{TP}$ count of the clause, while ``penalizing'' a clause refers to increasing its $\mathit{FP}$ or $\mathit{FN}$ counts. Penalizing clauses reduces their score, it therefore contributes to their specialization after a sufficient number of examples.}  
\label{fig:oled_init_term}
\end{figure}

To calculate $G$-scores, each clause $r$ is equipped with a true positive ($\mathit{TP}$), a false positive ($\mathit{FP}$) and a false negative ($\mathit{FN}$) counter, whose values are updated accordingly as $r$ gets evaluated on training examples that stream-in. True negative counts are not taken into account, since the annotation for complex events is acquired via the closed world assumption. Also, $r$ has an example counter that counts the number of examples on which $r$ has been evaluated so far and is used in the calculation of $\epsilon$ in the Hoeffding bound-based search heuristic. Although different scoring functions may be plugged into \oled, 
 in this work we use precision, to score initiation clauses, and recall, to score termination clauses, as in \cite{DBLP:journals/tplp/KatzourisAP16}. Moreover,  
\oled \ supports a clause pruning mechanism, that allows to remove low-quality clauses (e.g. clauses that have been generated from noisy examples) and a tie-breaking mechanism, that allows to randomly select between equally good specializations. We refer to \cite{DBLP:journals/tplp/KatzourisAP16} for more details on these features.   

In the general case, a theory learnt by \oled \ is a collection of clauses constructed with the online mechanism described above. Starting with an empty theory $H = \emptyset$, an initial clause is generated from the first positive example that streams-in, by constructing a bottom clause $\bot$ from that example and adding the empty-bodied clause $r = head(\bot) \leftarrow$ to theory $H$. From that point on, $r$ is gradually specialized by the addition of literals from $\bot$ to its body. New clauses are added to $H$ whenever existing clauses in $H$ become too specific to account for new incoming examples. 

When learning domain-specific axioms in the Event Calculus, the aforementioned generic theory construction strategy must be modified to account for the fact that initiation and termination clauses behave differently w.r.t. encountered $\mathit{TP, FP}$ and $\mathit{FN}$ complex event instances. A description of this behavior is illustrated in Figure \ref{fig:oled_init_term}(A). To handle this behavior, initiation and termination clauses are learnt separately, by two parallel processes, each of which runs the core \oled \ Algorithm. The input stream is forwarded to each of these processes simultaneously. Figure \ref{fig:oled_init_term}(B) presents the different actions that each learner takes whenever it encounters $\mathit{TP, FP}$ and $\mathit{FN}$ instances. 

\section{A Distributed Version of \oled}
\label{sec:d-oled}

We now proceed to the description of a distributed version of \oled. Evaluating a clause and its candidate specializations on incoming examples, may be performed in parallel, by distributing the clause evaluation workload across multiple processing nodes that operate on independent data partitions. When needed, e.g. when the Hoeffding test succeeds at some processing node, evaluation results from other nodes may be combined in order to make a more informed decision. We next describe this strategy in more detail.

We assume that learning is performed by a set $\mathcal{N}$ of independent, possibly distributed processing nodes. Each node $N_{i}\in \mathcal{N}$ handles a separate stream $S_{i}$ of training examples. The nodes in $\mathcal{N}$ communicate by exchanging messages and they learn a theory $H$ simultaneously, each node using its own training stream. The distributed version of \oled \ differs from the sequential algorithm in the following respects:

\textbf{New clause addition:} When a node $N_i$ generates a new clause $r$, it broadcasts $r$ to all other nodes in $\mathcal{N}$, via a $\mathit{\mathtt{AddNewClause}(r)}$ message (see Table \ref{table:messages} for the main types of message of distributed \oled). Each node that receives such a message adds clause $r$ to its own theory and starts scoring $r$, and its candidate specializations, on its own data. As in the single-core version of \oled, a new clause $r$ consists of an empty-bodied clause $\mathit{head(\bot_r)\leftarrow}$, where $\bot_r$ is a bottom clause generated at $N_i$. 

\textbf{Clause specialization:} When a node $N_i$ is about to specialize a clause $r$, i.e. when \oled's Hoeffding test for clause $r$ succeeds, locally at $N_i$, node $N_i$ sends a \linebreak $\mathit{\mathtt{SpecializeRequest}(r_{id})}$ message to all other nodes, where $r_{id}$ is a unique identifier of clause $r$, common to all copies of $r$ shared among processing nodes. Upon receiving such a message, each node uses $r_{id}$ to retrieve its own evaluation statistics for clause $r$ and its candidate specializations, which are sent over to the requesting node $N_i$. These statistics consist of $\mathit{TP, FP, FN}$ and $E$ counts for clause $r$ and its candidate specializations, where by $E$ we denote the number of examples on which a clause has been evaluated so far. The received counts for clause $r$ and its specializations are combined with node $N_i$'s local counts as follows (we describe the process for clause $r$ only, but it is similar for each one of its specializations). Denoting by $\mathit{TP_r^j, FP_r^j, FN_r^j}$ and $E_r^{j}$ the respective counts for clause $r$, received from node $N_j \in \mathcal{N}, j\neq i$, the current node $N_i$ updates $r$'s counts accordingly, by increasing $r$'s local counts with those received from other nodes. For instance, the new $\mathit{TP}$ count for clause $r$ in node $N_i$ becomes $\mathit{TP_r^i = TP_r^i + \sum\limits_{N_j\in \mathcal{N}}{TP_r^j}}$. $\mathit{FP_r^i, FN_r^i}$ and $E_r^i$ counts are updated in a similar fashion. 

Each processing node $N_i\in \mathcal{N}$ maintains a record, for each clause $r$ in its theory and each one of $r$'s specializations, that contains the exact counts previously received for them, from each node $N_j\in \mathcal{N}, j \neq i$. When node $N_i$ receives a set of new $\mathit{TP_r^j, FP_r^j, FN_r^j}$ and $E_r^j$ counts for clause $r$ from node $N_j, j\neq i$, the respective previous counts are subtracted from the new ones, to avoid over-scoring $r$ with counts that have already been taken into account in previous updates. The same holds for $r$'s specializations.

Once individual clause evaluation statistics are combined as described above, node $N_i$ repeats the Hoeffding test for clause $r$ to assess if the test still succeeds after the accumulated counts from all other nodes, for clause $r$ and its specializations, have been taken into account. If it does, clause $r$ is replaced in $H$, the current theory at node $N_i$, by its best-scoring specialization $r'$ that results from the Hoeffding test. Then node $N_i$ sends out a $\mathit{\mathtt{Replace}(r_{id},r')}$ message to all other nodes, instructing them to also replace $r$ in their own theories with $r'$. If, on the other hand, the Hoeffding test fails at node $N_i$ after the updated counts are taken into account, clause $r$ is not specialized and all nodes continue evaluating their theories on new incoming examples from their training streams.

\begin{table}[t]
\begin{tabularx}{\linewidth}{*{3}{C}}
\toprule
\textbf{Message} & \textbf{Conditions for meassage broadcast} & \textbf{Actions upon message receipt}\\\midrule
$\mathit{\mathtt{AddNewClause}(r)}$ & Generation of clause $r$. &  Add $r$ to local theory.\\
\bottomrule
$\mathit{\mathtt{SpecializeRequest}(r_{id})}$ & Clause with id $r_{id}$ is about to be specialized. & Reply to the sender by the local $\mathit{TP, FP, FN, E}$ counts for clause with id $r_{id}$ and for each of its candidate specializations. \\
\bottomrule
$\mathit{\mathtt{Replace}(r_{id},r')}$ & Clause with id $r_{id}$ has been specialized to clause $r'$. & Replace clause with id $r_{id}$ by $r'$ in local theory. \\
\bottomrule
$\mathit{\mathtt{PruneRequest}(r_{id})}$ & Clause with id $r_{id}$ is about to be pruned. & Reply to the sender by the local $\mathit{TP, FP, FN, E}$ counts for clause with id $r_{id}$, as well as the period for which $r$ remains (locally) unchanged.\\
\bottomrule
$\mathit{\mathtt{Remove}(r_{id})}$ & Clause with id $r_{id}$ has been pruned. & Remove clause with id $r_{id}$ from local theory.  \\
\bottomrule
\end{tabularx}
\vspace*{0.01cm}
\caption{The main messages exchanged between data processing nodes in distributed \oled. }\label{table:messages}
\end{table}

\textbf{Clause pruning:} 
For a clause $r$ to be pruned away, two conditions must hold: 
 First, clause $r$ must be unchanged (not specialized) for a sufficiently long period, which, in the single-core version of \oled, is set to the average number of examples, observed so far in the learning process, for which the Hoeffding test succeeds, i.e. the average  value 
 of $n = \mathcal{O}(\frac{1}{\epsilon^2}ln\frac{1}{\delta})$ that has resulted in clause specializations so far. Second, from that point on where clause $r$ remains unchanged, a sufficiently large number of examples must be seen, in order to use a Hoeffding test to infer that, with probability $1 - \delta$, the quality of clause $r$ is below the pruning threshold.

\begin{algorithm}[t]
\small
 \caption{\textsf{\small OLEDNode}$(H, G, \mathcal{I},\mathcal{N}')$\newline
 \small
\textbf{Input: }$H$: A potentially empty hypothesis; $G$: A clause evaluation function; $\mathcal{I}$: A stream of training interpretations; $\mathcal{N}'$: Set of peer processing nodes.
\normalsize}
 \label{alg:d-oled} 
\begin{algorithmic}[1]
\State $H := \emptyset$ \label{alg:main_start}
\While{\texttt{true}}
\State \textbf{Let} $I$ be the next training interpretation
\State \begin{varwidth}[t]{\linewidth} $G$-score each clause $r\in H$ on $I$  \end{varwidth}\label{alg:update_statistics}
  \If{a new clause should be generated}\label{alg:expand_theory_test}
    \State Generate a new clause $r$ and add to $H$.
    \State Send $r$ to each node in $\mathcal{N}'$. 
  \EndIf
    \ForAll{clause $r \in H$}
      \If{the Hoeffding test for $r$ succeeds}
        \State Request the counts for $r$, and all of $r$'s specializations, from all nodes in $\mathcal{N}'$.
        \State \begin{varwidth}[t]{\linewidth}Add the received counts to the current ones and repeat the Hoeffding test. 
        \end{varwidth} 
        \If{the Hoeffding test for $r$ still succeeds}
         \State Replace $r$ in $H$ with its best-scoring specialization. 
         \State Notify all nodes in $\mathcal{N}'$ to also replace $r$ with its best-scoring specialization.        
        \EndIf
      \EndIf
      \If{$r$ should be pruned}
        \State Request the counts for $r$ from all nodes in $\mathcal{N}'$.
        \State Add the received counts for $r$ to the current ones.
        \If{$r$ still should be pruned}
        \State Remove $r$ from the current theory.
        \State Notify all nodes in $\mathcal{N}'$ to also remove $r$ from their theories.      
        \EndIf      
      \EndIf
    \EndFor  
\EndWhile
\State \textbf{return} $H$ \label{alg:main_end}
\end{algorithmic}
\end{algorithm}
\normalsize

In the distributed version of \oled, each node uses the above heuristics to decide locally whether a clause $r$ should be pruned. Once it has seen enough data from its own stream to make that decision for clause $r$, it sends a $\mathit{\mathtt{PruneRequest}(r_{id})}$ message to all other nodes. Each node that receives such a message sends back to the requesting node the necessary information (period for which clause $r$ remains unchanged and $\mathit{TP, FP, FN}$ and $E_r$ counts for clause $r$), which node $N_i$ uses to re-assess whether clause $r$ should be pruned, based on the global view of clause $r$, obtained by combining $r$'s separate evaluations from all processing nodes. If node $N_i$ eventually decides to pruned clause $r$, it sends a $\mathit{\mathtt{Remove}(r_{id})}$ to all other nodes, which instructs them to also remove clause $r$ from their theories. 

Algorithm \ref{alg:d-oled} illustrates learning in distributed \oled \  by displaying the functionality of each processing node.

To ensure that all nodes have the same theory at each point during the learning process, processing nodes often block their execution. For instance, whenever a node $N_j$ sends out  a $\mathit{\mathtt{SpecializeRequest}(r_{id})}$, it blocks untils it receives the necessary statistics for clause $r_{id}$ from all other nodes. Similarly, whenever a node $N_i, i\neq j$ receives such a message, it replies by sending over to $N_j$ the necessary statistics for clause $r_{id}$ and then blocks its execution, waiting for a ``verdict'' from $N_j$. A ``verdict'' may be a \texttt{Replace}$(r_{id}, r')$ message, which instructs node $N_i$ to replace clause $r_{id}$ by its specialization $r'$, or, it may be a $\mathit{\mathtt{Proceed}}$ message (this type of message is omitted from Table \ref{table:messages}), signifying the fact that the Hoeffding test for clause $r_{id}$ at node $N_j$ failed, after accumulated counts from all other nodes were taken into account, and therefore node $N_i$ may continue processing new examples without altering its current theory. A similar blocking behavior occurs during message passing for clause pruning, to ensure that all all nodes remove (or preserve) a clause in a synchronized fashion. 

While exchanging messages during learning, care must be taken to avoid deadlocks and race conditions. An example of such situations is the case where the Hoeffding test for specializing a single clause $r$ succeeds simultaneously at two different nodes $N_i$ and $N_j$. In such a case, nodes $N_i$ and $N_j$ send a  $\mathit{\mathtt{SpecializeRequest}(r_{id})}$ message to each other (in addition to all other nodes), requesting each other's counts for clause $r$. Subsequently, both nodes enter a ``waiting state'' as mentioned above, waiting to receive the requested counts in order to proceed with repeating the Hoeffding test. This results in a deadlock, since each node is waiting for the other node's reply. A similar situation may occur with pruning. To avoid such problems, all communication between nodes regarding clause specialization and pruning is mediated by a separate node, whose sole role is to assign priorities in such cases, so that such deadlocks are avoided. When two or more nodes request to specialize or prune a clause (almost) simultaneously, the mediator node prioritizes one of them randomly, enqueueing the other nodes for proceeding at a later time. The enqueued nodes reply to the requests of the prioritized node. Once the prioritized node finishes, the next node from the queue is prioritized. In such cases, a node from the queue abandons its effort to specialize or prune a clause $r$, if $r$ gets specialized or pruned by the previously prioritized node.  
  
To sum-up, all nodes in $\mathcal{N}$ share a copy of the same theory $H$ at each point in learning. $H$ is learnt in an online fashion, simultaneously from all nodes in $\mathcal{N}$, each node handling its own training stream.  
As in the single-core version, initiation and termination clauses are learnt independently, by two separate groups of processes, each of which learns one type of clause and implements the distributed version of \oled \ described in this section. 

\section{Experimental Evaluation}
\label{sec:experiments}

\begin{table}[t]
\begin{tabularx}{1\linewidth}{*{9}{C}}
\toprule
~ & ~ & \textbf{$\#$cores} & \textbf{Time (sec)} & \textbf{Speed-up} & \textbf{$F_1$-score} & \textbf{Theory size} & \textbf{$\#$Msgs} & \textbf{Msg size (KB)}\\ \midrule

\textbf{(A)} & \emph{Meeting} & 1 & 46 & -- & 0.794 & \textbf{28} & --  & --\\
~ & ~ & 2 & 17 & 2.7 & \textbf{0.802} & 32 & \textbf{34}  & \textbf{97}\\
~ & ~ & 4 &  15 & 3  & 0.798 & 32 & 52  & 122  \\
~ & ~ & 8 &  \textbf{14} & \textbf{3.2} & 0.800 & 34 & 88  & 182\\

\noalign{\smallskip}

~ & \emph{Moving} & 1  & 73 & -- & \textbf{0.731} & \textbf{21} & --  & -- \\
 & ~ & 2 &  34 & 2 & 0.729 & 21 & \textbf{41}  & \textbf{102}\\
~ & ~ & 4  &  30 & 2.4 & \textbf{0.731} & 23 & 64  & 188   \\
~ & ~ & 8 & \textbf{28} & \textbf{2.6} & 0.730 & 23 & 108  & 218\\

\hline\noalign{\smallskip}

\textbf{(B)} & \emph{Meeting} & 1 & 423 & -- & 0.832 & \textbf{36} & --  & --\\
~ & ~ & 2 & 122 & 3.4 & 0.832 & 36 & \textbf{105}  & \textbf{247}\\
~ & ~ & 4 &  116 & 3.6  & 0.832 & 36 & 298  & 705  \\
~ & ~ & 8 &  \textbf{110} & \textbf{3.8} & 0.832 & 36 & 518  & 1527\\

\noalign{\smallskip}

~ & \emph{Moving} & 1  & 514 & -- & 0.758 & \textbf{21} & --  & -- \\
 & ~ & 2 &  238 & 2.1 & 0.758 & 21 & \textbf{97}  & \textbf{218}\\
~ & ~ & 4  &  225 & 2.2 & 0.758 & 21 & 265  & 607   \\
~ & ~ & 8 & \textbf{221} & \textbf{2.3} & 0.758 & 21 & 482 & 1422  \\

\bottomrule
\end{tabularx}
\vspace*{0.01cm}
\caption{\footnotesize \textbf{(A)} Experimental results from the CAVIAR dataset; \textbf{(B)} Experimental results from a dataset consisting of 10 copies of CAVIAR.}\label{table:results}
\end{table}

We present an experimental evaluation of our approach on CAVIAR (described in Section \ref{sec:backgound}), a benchmark dataset for activity recognition. CAVIAR contains 282,067 training interpretations with a mean size of 25 atoms each.
  \oled \ is implemented in the Scala programming language. It uses \textsf{\small Clingo}\footnote{\url{http://potassco.sourceforge.net/}} as its main reasoning component and Scala's \textsf{\small akka} Actors library\footnote{\url{http://akka.io/}} to model the behavior of a processing node (Algorithm \ref{alg:d-oled}) and implement message passing. The code and data are available online\footnote{\url{https://github.com/nkatzz/OLED}}.  All experiments were conducted on a Linux machine with a 3.6GHz processor (4 cores and 8 threads) and 16GB of RAM. 

The purpose of our experiments was to compare the distributed version of \oled \ with its monolithic counterpart. We performed learning with 1, 2, 4 and 8 processing cores for acquiring the definitions of two target complex events, related to two persons \emph{meeting each other} or \emph{moving together}. CAVIAR contains  6,272 interpretations in which \emph{moving} occurs and 3,722 in which \emph{meeting} occurs. For the experiments with the distributed version of \oled \ positive and negative interpretations for the target complex events were evenly distributed across different processing cores. The results 
 were obtained by tenfold cross-validation  and are presented in Table \ref{table:results}(A), in the form of averages 
for training time, $F_1$-score and theory size (total number of literals), as well as average exchanged message number and size. 
 $F_1$-scores were obtained by micro-averaging results from each fold.

With respect to the predictive accuracy, multiple-core learning resulted in theories of slightly higher $F_1$-score for the \emph{meeting} complex event, as compared to single-core learning. In the monolithic setting, \oled \ postpones the generation of new clauses, up to the point where existing clauses become too specific to cover new examples. During this time, positive examples which may result in good clauses (recall that \oled \ learns by ``encoding'' examples into bottom clauses), are ``skipped'', i.e. they are not used for learning new clauses, since they are covered by existing ones. In contrast, the data distribution in the multi-core  setting results in cases where interesting examples that would have been missed 
in the monolithic setting, are actually used for learning. This resulted in \oled \ learning slightly ``richer'' theories for the \emph{meeting} complex event in the multi-core setting.
 A similar effect was not observed for the \emph{moving} complex event, which has a simpler definition than \emph{meeting}.


Regarding training times, OLED achieves a signicant speed-up, by moving from monolithic learning to learning with two cores. This is achieved by exchanging very little information between the processing cores (see the last column of Table \ref{table:results}(A)). When additional cores are added, the speed-up is sub-linear, mainly due to the increased number of exchanged messages that result in blocking processing cores.

To test distributed OLED further, we used a larger dataset obtained by sequentially appending to each other 10 copies of CAVIAR, ``pushing forward'' the time-stamps of the data in each copy. 
 The experimental setting for the $\times$10 CAVIAR experiment was identical to the one described above for the $\times$1 CAVIAR experiment. The results are presented in Table \ref{table:results}(B).

Training times in the $\times$10 CAVIAR experiment follow the same speed-up pattern observed in the $\times$1 experiment. In this set of experiments, the speed-up factor reaches 3.8, leading to highly reduced training times.

Due to the repetition of the training data in the $\times$10 CAVIAR experiment, each fold of the tenfold cross-validation process converged to identical theories, regardless of the number of cores used. $F_1$-scores are therefore identical for all number of cores, and are also improved as compared to the $\times$1 CAVIAR experiment. In the latter experiment, good clauses were often constructed ``too-late'', from examples that were encountered shortly before the data were exhausted (in which case \oled \ terminates). Such clauses 
 may be discarded, since \oled \ uses a ``warm-up'' period  parameter that controls a minimum number of examples on which a clause must be evaluated in order to be included in an output hypothesis. 
 In contrast, in the $\times$10 CAVIAR experiment such problems were avoided, thanks to the increase in training data size. 


\section{Related Work}
\label{sec:related-work}
An overview of existing approaches to learning theories in the Event Calculus with ILP may be found at \cite{katzouris2015incremental,katzourisPhd} and a discussion on how   \oled \ compares to such approaches may be found at \cite{DBLP:journals/tplp/KatzourisAP16,katzourisPhd}. In this section we mainly discuss distributed/parallel ILP algorithms, for which a substantial amount of work exists in the literature. A thorough review may be found in \cite{fonseca2009parallel,srinivasan2012data}. Parallel ILP algorithms exploit parallelism across three main axes \cite{fonseca2009parallel}: Searching through
the hypothesis space in parallel (search parallelism); splitting the training data and learning from data subsets (data parallelism); and evaluating candidate clauses in parallel (evaluation/coverage parallelism). 

In \cite{skillicorn2001parallel} the authors present a a data-parallel version of a standard set-cover loop: Each processing node learns a fragment of the concept definition from a partition of the data, and then these fragments are exchanged between all nodes. Good-enough clauses are kept by all nodes. A cover removal step is subsequently implemented by each core and the set-cover loop continues. Overall, the approach in \cite{skillicorn2001parallel} learns much faster that a sequential algorithm, achieving super-linear speed-ups. A similar approach is proposed in \cite{DBLP:conf/cluster/FonsecaSCC05}, where the training examples are split across multiple nodes and searched in parallel, while the best rules from each node are ``pipe-lined'' to all other nodes. 

In \cite{srinivasan2012data} the authors use a MapReduce-based framework to parallelize the operation of a classical set-cover ILP algorithm towards both evaluation-parallelism and search-parallelism. In the former case, coverage tests of candidate clauses are performed in parallel, on disjoint partitions of the data. In the latter case, bottom clauses (which are generalized to acquire a hypothesis clause) are generated and searched in a concurrent fashion from more than one ``seed'' examples. The reducer then selects the best hypothesis clause that results from this process. A similar approach for parallel exploration of independent hypotheses has been proposed in \cite{ohwada1999parallel}, while similar approaches towards parallel coverage tests have been proposed in \cite{DBLP:conf/csb/GrahamPK03,fidjeland2014customisable}. In \cite{nishiyama2015yet}, the approach of \cite{srinivasan2012data} was extended to a framework that is capable of self-regulating the workload of distributing learning costs across multiple nodes. 

A main difference of the work presented here from the aforementioned approaches to parallel ILP is that they all rely on iterative ILP algorithms (basically, a set-cover loop), and they all require several passes over the data to compute a hypothesis. In contrast, \oled \ is an online, single-pass algorithm. In relation to the latter, some work on streaming ILP exists. However, existing approaches are either oriented towards unsupervised tasks like frequent pattern discovery \cite{silva2015multi}, or they rely on propositionalization techniques and off-the-self, online propositional learners \cite{srinivasan2013relational}.  


\section{Conclusions and Future Work}
\label{sec:final}

We presented a distributed version of a recently proposed algorithm for online learning of complex event definitions in the form of domain-specific axioms in the Event Calculus. We also presented an experimental evaluation of our approach on a benchmark dataset for activity recognition, which demonstrates that we can significantly reduce training times. As future work, we aim to evaluate our approach on larger datasets, in terms of in-situ, geographically distributed learning, as in the case of maritime monitoring. 
 Also, we plan to 
 remove the requirement that all processing nodes block their execution, while waiting for replies during message passing.

\subsubsection*{Acknowledgments.} This work was funded by the H2020 project DATACRON.

\bibliographystyle{plain}
\bibliography{refs}

\begin{thebibliography}{10}

\bibitem{artikis2015event}
Alexander Artikis, Marek Sergot, and Georgios Paliouras.
\newblock An event calculus for event recognition.
\newblock {\em Knowledge and Data Engineering, IEEE Transactions on},
  27(4):895--908, 2015.

\bibitem{blockeel1998top}
Hendrik Blockeel and Luc De~Raedt.
\newblock Top-down induction of first-order logical decision trees.
\newblock {\em Artificial intelligence}, 101(1):285--297, 1998.

\bibitem{blockeel1999scaling}
Hendrik Blockeel, Luc De~Raedt, Nico Jacobs, and Bart Demoen.
\newblock Scaling up inductive logic programming by learning from
  interpretations.
\newblock {\em Data Mining and Knowledge Discovery}, 3(1):59--93, 1999.

\bibitem{de2008logical}
Luc De~Raedt.
\newblock {\em Logical and relational learning}.
\newblock Springer Science \& Business Media, 2008.

\bibitem{etzion2010event}
Opher Etzion and Peter Niblett.
\newblock {\em Event processing in action}.
\newblock Manning Publications Co., 2010.

\bibitem{fidjeland2014customisable}
Andreas~K Fidjeland, Wayne Luk, and Stephen~H Muggleton.
\newblock Customisable multi-processor acceleration of inductive logic
  programming.
\newblock {\em Latest Advances in Inductive Logic Programming}, pages 123--141,
  2014.

\bibitem{DBLP:conf/cluster/FonsecaSCC05}
Nuno~A. Fonseca, Fernando M.~A. Silva, V{\'{\i}}tor~Santos Costa, and Rui
  Camacho.
\newblock A pipelined data-parallel algorithm for {ILP}.
\newblock In {\em 2005 {IEEE} International Conference on Cluster Computing
  {(CLUSTER} 2005), September 26 - 30, 2005, Boston, Massachusetts, {USA}},
  pages 1--10, 2005.

\bibitem{fonseca2009parallel}
Nuno~A Fonseca, Ashwin Srinivasan, Fernando Silva, and Rui Camacho.
\newblock Parallel ilp for distributed-memory architectures.
\newblock {\em Machine learning}, 74(3):257--279, 2009.

\bibitem{gama2010knowledge}
Joao Gama.
\newblock {\em Knowledge discovery from data streams}.
\newblock CRC Press, 2010.

\bibitem{gebser2012answer}
Martin Gebser, Roland Kaminski, Benjamin Kaufmann, and Torsten Schaub.
\newblock Answer set solving in practice.
\newblock {\em Synthesis Lectures on Artificial Intelligence and Machine
  Learning}, 6(3):1--238, 2012.

\bibitem{DBLP:conf/csb/GrahamPK03}
James~H. Graham, C.~David~Page Jr., and Ahmed~H. Kamal.
\newblock Accelerating the drug design process through parallel inductive logic
  programming data mining.
\newblock In {\em 2nd {IEEE} Computer Society Bioinformatics Conference, {CSB}
  2003, Stanford, CA, USA, August 11-14, 2003}, pages 400--402, 2003.

\bibitem{hoeffding1963probability}
Wassily Hoeffding.
\newblock Probability inequalities for sums of bounded random variables.
\newblock {\em Journal of the American statistical association},
  58(301):13--30, 1963.

\bibitem{katzourisPhd}
Nikos Katzouris.
\newblock Scalable relational learning for event recognition.
\newblock {\em PhD Thesis, University of Athens}, 2017.

\bibitem{katzouris2015incremental}
Nikos Katzouris, Alexander Artikis, and Georgios Paliouras.
\newblock Incremental learning of event definitions with inductive logic
  programming.
\newblock {\em Machine Learning}, 100(2-3):555--585, 2015.

\bibitem{DBLP:journals/tplp/KatzourisAP16}
Nikos Katzouris, Alexander Artikis, and Georgios Paliouras.
\newblock Online learning of event definitions.
\newblock {\em {TPLP}}, 16(5-6):817--833, 2016.

\bibitem{kowalski1986logic}
Robert Kowalski and Marek Sergot.
\newblock A logic-based calculus of events.
\newblock {\em New Generation Computing}, 4(1):67--95, 1986.

\bibitem{nishiyama2015yet}
Hiroyuki Nishiyama and Hayato Ohwada.
\newblock Yet another parallel hypothesis search for inverse entailment.
\newblock In {\em ILP}, 2015.

\bibitem{ohwada1999parallel}
Hayato Ohwada and Fumio Mizoguchi.
\newblock Parallel execution for speeding up inductive logic programming
  systems.
\newblock In {\em International Conference on Discovery Science}, pages
  277--286. Springer, 1999.

\bibitem{silva2015multi}
Andreia Silva and Cl{\'a}udia Antunes.
\newblock Multi-relational pattern mining over data streams.
\newblock {\em Data Mining and Knowledge Discovery}, 29(6):1783--1814, 2015.

\bibitem{skillicorn2001parallel}
David~B Skillicorn and Yu~Wang.
\newblock Parallel and sequential algorithms for data mining using inductive
  logic.
\newblock {\em Knowledge and Information Systems}, 3(4):405--421, 2001.

\bibitem{srinivasan2013relational}
Ashwin Srinivasan and Michael Bain.
\newblock Relational models with streaming ilp.
\newblock In {\em ILP}, 2013.

\bibitem{srinivasan2012data}
Ashwin Srinivasan, Tanveer~A Faruquie, and Sachindra Joshi.
\newblock Data and task parallelism in ilp using mapreduce.
\newblock {\em Machine learning}, 86(1):141--168, 2012.

\end{thebibliography}

\end{document}